\documentclass[letterpaper,12pt,notitlepage]{report}
\usepackage[dvips]{epsfig}
\usepackage{color}
\usepackage{float}
\usepackage[centertags]{amsmath}
\usepackage{amsfonts}
\usepackage{amssymb}
\usepackage{algorithm}
\usepackage[noend]{algorithmic}
\usepackage{graphics}
\usepackage{graphicx}
\usepackage{amsthm}
\usepackage{subfigure}
\usepackage{newlfont}
\usepackage{upgreek}
\usepackage[latin1]{inputenc}
\usepackage{sty/utfsm_tesis}
\usepackage{sty/xtocinc} 
\newlength{\defbaselineskip}
\newcommand{\setlinespacing}[1]%
	   {\setlength{\baselineskip}{#1 \defbaselineskip}}

\newcommand{\singlespacing}{\setlength{\baselineskip}{\defbaselineskip}}

\theoremstyle{plain}

\theoremstyle{definition}

\theoremstyle{remark}

\numberwithin{equation}{section}

\begin{document}


\dedicate{Real stupidity beats artificial intelligence every time.\\
\hfill Terry Pratchett}
\magister

\copyrightyear{2009} \submitdate{Nov 2009}
\convocation{November}{2009}


\title{Single-Agent On-line Path Planning in Continuous, Unpredictable and Highly Dynamic Environments}

\author{Nicol\'as Arturo Barriga Richards}


\profguia{Dr. Mauricio Solar}

\profext{Dr. John Atkinson}

\profcorr{Dr. Horst H. von Brand}

 \ack{tex/Acknowledgements}			   
 \resumenesp{tex/Resumen}	      
 \resumening{tex/Abstract}		   
 \abreviaciones{tex/Abbreviations}   

\beforepreface
\afterpreface

\numberwithin{equation}{chapter}
\chapter{Introduction}
The \emph{dynamic path-planning} problem consists in finding a suitable
plan for each new configuration of the environment by recomputing a
collision-free path using the new information available at each time step~\cite{Hwang92}.
This kind of problem has to be solved for example by a robot trying to navigate
through an area crowded with people, such as a shopping mall or supermarket.
The problem has been widely addressed in its several flavors,
such as cellular decomposition of the configuration space~\cite{Stentz95},
partial environmental knowledge~\cite{Stentz94},
high-dimensional configuration spaces~\cite{Kavraki96}
or planning with non-holonomic constraints~\cite{Lavalle99}.
However, even simpler variations of this problem are complex enough
that they can not be solved
with deterministic techniques, and therefore are worthy of study.

This thesis is focused on algorithms for
finding and traversing a collision-free path in
two dimensional space, for a
holonomic robot\footnote{A holonomic robot is a robot in
which the controllable degrees of freedom is equal to the total degrees of
freedom.},
without kinodynamic restrictions\footnote{Kinodynamic planning is a problem in which velocity and acceleration
bounds must be satisfied}, in a highly dynamic environment, but for comparison
purposes three different scenarios will be tested:
\begin{itemize}
\item Several unpredictably moving obstacles or adversaries.
\item Partially known environment, where some obstacles become visible when the
robot approaches each one of them.
\item Totally unknown environment, where every obstacle is initially invisible
to the planner, and only becomes visible when the robot approaches it.
\end{itemize}
Besides the obstacles in the second and third scenario we assume
that we have perfect information of the environment at all times.

We will focus on continuous space algorithms and
will not consider algorithms that use a discretized representation of the
configuration space\footnote{the space of possible positions that a physical
system may attain}, such
as D*~\cite{Stentz95},
because for high dimensional problems the
configuration space becomes intractable in terms of both memory and computation
time, and there is the extra difficulty of calculating the discretization size,
trading off accuracy versus computational cost.
Only single agent algorithms will be considered here.
On-line as well as off-line algorithms will be studied. An on-line algorithm is
one that is permanently adjusting its solution as the environment changes, while
an off-line algorithm computes a solution only once (however, it can be executed 
many times).

The offline Rapidly-Exploring Random Tree (RRT) is efficient at finding solutions, but the results are far from
optimal, and must be post-processed for shortening, smoothing or other qualities
that might be desirable in each particular problem. Furthermore, replanning RRTs
are costly in terms of computation time, as are evolutionary and cell-decomposition
approaches. Therefore, the novelty of this work is
the mixture of the feasibility benefits of the RRTs, the repairing capabilities
of local search, and the computational inexpensiveness of greedy algorithms,
into our lightweight multi-stage algorithm. Our working hypothesis will be that
a multi-stage algorithm, using different techniques for initial planning and
navigation,
outperforms current probabilistic sampling techniques in highly dynamic
environments

\section{Problem Formulation}

At each time-step, the problem could be defined as
an optimization problem with satisfiability constraints.
Therefore, given a path our objective is to minimize an evaluation function
(i.e., distance, time, or path-points), with the \(C_{\text{free}}\) constraint.
Formally, let the path \mbox{$\rho=p_1p_2\ldots p_n$} be a sequence of points, where
\mbox{$p_i \in \mathbb{R}^n$} is a $n$-dimensional point (\mbox{\(p_1 = q_{\text{init}}, p_n =
q_{\text{goal}}\)}),
$O_t\in \mathcal{O} $ the set of obstacles positions
at time $t$, and \mbox{$\operatorname{eval} \colon \mathbb{R}^n \times
\mathcal{O} \mapsto \mathbb{R}$}
an evaluation function of the path depending on the object positions.
Our ideal objective is to obtain the optimum $\rho^*$ path that
minimizes our $\operatorname{eval}$ function within a feasibility restriction in the form
\begin{equation}
\displaystyle\rho^*=\underset{\rho}{\operatorname{argmin}}[\operatorname{eval}(\rho,O_t)]  \textrm{ with }
\operatorname{feas}(\rho,O_t) = C_{\text{free}}
\label{eq:problem}
\end{equation}
where \(\operatorname{feas}(\cdot,\cdot)\) is a \emph{feasibility} function that equals
\(C_{\text{free}}\)
if the path $\rho$ is collision free for the obstacles $O_t$.
For simplicity, we use very naive \(\operatorname{eval}(\cdot,\cdot)\) and
\(\operatorname{feas}(\cdot,\cdot)\)
functions, but our approach could be extended easily to more complex evaluation and
feasibility functions.
The \(\operatorname{feas}(\rho,O_t)\) function used assumes that the robot is a point
object in space, and therefore if no segments
\(\overrightarrow{p_i p_{i+1}}\)
of the path collide with any object \(o_j \in O_t\), we say that the path
is in \(C_{\text{free}}\).
The \(\operatorname{eval}(\rho,O_t)\) function is the length of the path, i.e., the sum of the distances between
consecutive points. This could be easily changed to any other metric such as the time
it would take to traverse this path, accounting for smoothness,
clearness or several other optimization criteria.

\section{Document Structure}

In the following sections we present several path planning methods that can be
applied to the problem described above. In section~\ref{sec:RRT} we review
the basic offline, single-query RRT, a probabilistic method that builds a
tree along the free configuration space until it reaches the goal state.
Afterwards, we introduce the most popular replanning variants of
RRT: Execution Extended RRT (ERRT) in section~\ref{sec:ERRT}, Dynamic RRT (DRRT) in
section~\ref{sec:DRRT} and Multipartite RRT (MP-RRT) in section~\ref{sec:MPRRT}.
The Evolutionary Planner/Navigator (EP/N), along with some variants, is presented
in section~\ref{sec:EP/N}.
Then, in section~\ref{sec:RRT-EP/N} we present a mixed approach, using a RRT to
find an initial solution and the EP/N to navigate, and finally, in section~\ref{sec:RRT-LP} we
present our new hybrid multi-stage algorithm, that uses RRT for initial
planning and informed local search for navigation, plus a simple greedy
heuristic for optimization. Experimental results and
comparisons that show that this combination of simple techniques provides better
responses to highly dynamic environments than the standard RRT extensions are
presented in section~\ref{sec:results}. The
conclusions and further work are discussed in section~\ref{sec:conclusions}.

\chapter{State of the Art}
\label{sec:stateoftheart}
In this chapter we present several path planning methods that can be
applied to the problem described above. First we will introduce variations of
the Rapidly-Exploring Random Tree (RRT), a probabilistic method that builds a
tree along the free configuration space until it reaches the goal state. This
family of planners is fast at finding solutions, but the solutions are far from
optimal, and must be post-processed for shortening, smoothing or other qualities
that might be desirable in each particular problem. Furthermore, replanning RRTs
are costly in terms of computation time. We then introduce an
evolutionary planner with somewhat opposite qualities: It is slow in finding
feasible solutions in difficult maps, but efficient at replanning when a
feasible solution has already been found. It can also optimize the solution
according to any given fitness function without the need for a post-processing
step.
\section{Rapidly-Exploring Random Tree}
\label{sec:RRT}
One of the most successful probabilistic methods for offline path planning
currently in use is the Rapidly-Exploring Random Tree (RRT), a single-query planner for
static environments, first introduced in~%
\cite{Lavalle98}. RRTs works towards finding a continuous path from a state
\(q_{\text{init}}\) to a state \(q_{\text{goal}}\) in the free configuration
space \(C_{\text{free}}\) by
building a tree rooted at \(q_{\text{init}}\). A new state \(q_{\text{rand}}\) is uniformly
sampled at random from the configuration space \(C\). Then the nearest node,
\(q_{\text{near}}\), in
the tree is located, and if \(q_{\text{rand}}\) and the shortest path from
\(q_{\text{rand}}\) to
\(q_{\text{near}}\) are in \(C_{\text{free}}\), then \(q_{\text{rand}}\) is added to the tree (algorithm~%
\ref{alg:buildrrt}). The tree
growth is stopped when a node is found near \(q_{\text{goal}}\). To speed up convergence,
the search is usually biased to \(q_{\text{goal}}\) with a small probability.

In~\cite{Kuffner00}, two new features are added to RRT. First, the EXTEND
function (algorithm~\ref{alg:extend}) is introduced, which instead of trying
to add \(q_{\text{rand}}\) directly to
the tree, makes a motion towards \(q_{\text{rand}}\) and tests for collisions.

\begin{algorithm}[ht!]
    \caption{$\operatorname{BuildRRT}(q_{\text{init}},q_{\text{goal}})$}
    \label{alg:buildrrt}
    \begin{algorithmic}[1]
        \STATE{\(T \leftarrow \text{empty tree}\)}
        \STATE \(T.\operatorname{init}(q_{\text{init}})\)
        \WHILE {\(\operatorname{Distance}(T,q_{\text{goal}})> \text{threshold}\)}
            \STATE \(q_{\text{rand}} \leftarrow \operatorname{RandomConfig}()\)
            \STATE \(\operatorname{Extend}(T,q_{\text{rand}})\)
        \ENDWHILE
        \RETURN \(T\)
    \end{algorithmic}
\end{algorithm}

\begin{figure}[ht!]
\begin{center}
\includegraphics[width=0.4\textwidth]{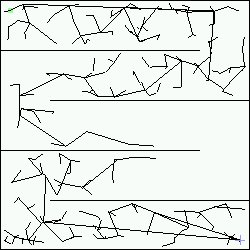}
\caption{RRT during execution}
\label{fig:rrt}
\end{center}
\end{figure}

\begin{algorithm}[ht!]
    \caption{$\operatorname{Extend}(T,q)$}
    \label{alg:extend}
    \begin{algorithmic}[1]
        \STATE \(q_{\text{near}} \leftarrow \operatorname{NearestNeighbor}(q,T)\)
        \IF{\(\operatorname{NewConfig}(q,q_{\text{near}},q_{\text{new}})\)}
            \STATE \(T.\operatorname{add\_vertex}(q_{\text{new}})\)
            \STATE \(T.\operatorname{add\_edge}(q_{\text{near}},q_{\text{new}})\)
            \IF{\(q_{\text{new}} = q\)}
                \RETURN Reached
            \ELSE
                \RETURN Advanced
            \ENDIF
        \ENDIF
        \RETURN Trapped
    \end{algorithmic}
\end{algorithm}
Then a
greedier approach is introduced (the CONNECT function, shown in algorithms~%
\ref{alg:rrtconnectplanner} and~%
\ref{alg:connect}), which repeats EXTEND until
an obstacle is reached. This ensures that most of the time we
will be adding states to the tree, instead of just rejecting new random states.
The second extension is the use of two trees, rooted at \(q_{\text{init}}\) and
\(q_{\text{goal}}\), which are grown towards each other (see figure~%
\ref{fig:rrt}). This significantly decreases the
time needed to find a path.

\begin{algorithm}[ht!]
    \caption{$\operatorname{RRTConnectPlanner}(q_{\text{init}},q_{\text{goal}})$}
    \label{alg:rrtconnectplanner}
    \begin{algorithmic}[1]
        \STATE \(T_a \leftarrow \text{tree rooted at \(q_{\text{init}}\)}\)
        \STATE \(T_b \leftarrow \text{tree rooted at \(q_{\text{goal}}\)}\)
        \STATE \(T_a.\operatorname{init}(q_{\text{init}})\)
        \STATE \(T_b.\operatorname{init}(q_{\text{goal}})\)
        \FOR{\(k=1\) to \(K\)}
            \STATE \(q_{\text{rand}} \leftarrow \operatorname{RandomConfig}()\)
            \IF{\textbf{not} (\(\operatorname{Extend}(T_a,q_{\text{rand}})=\text{Trapped}\))}
                \IF{\(\operatorname{Connect}(T_b,q_{\text{new}})=\text{Reached}\)}
                    \RETURN \(\operatorname{Path}(T_a,T_b)\)
                \ENDIF
            \ENDIF
            \STATE \(\operatorname{Swap}(T_a,T_b)\)
        \ENDFOR
        \RETURN Failure
    \end{algorithmic}
\end{algorithm}

\begin{algorithm}[ht!]
    \caption{$\operatorname{Connect}(T,q)$}
    \label{alg:connect}
    \begin{algorithmic}[1]
        \REPEAT
            \STATE \(S \leftarrow \operatorname{Extend}(T,q)\)
        \UNTIL{\((S \ne \text{Advanced})\)}
    \end{algorithmic}
\end{algorithm}

\section{Retraction-Based RRT Planner}
\label{sec:RRRT}
The Retraction-based RRT Planner presented in~\cite{Zhang08} aims at
improving the performance of the standard offline RRT in static
environments with narrow passages. The basic idea of the
\(\operatorname{Optimize}(q_r,q_n)\)
function in algorithm~%
\ref{alg:rrrt} is to iteratively retract a randomly generated
configuration
that is in \(C_{\text{obs}}\) to the closest boundary point in
\(C_{\text{free}}\). So, instead of
using the standard extension that tries to extend in a straight line from
\(q_{\text{near}}\) to \(q_{\text{rand}}\), it extends from \(q_{\text{near}}\) to the closest point in
\(C_{\text{free}}\) to \(q_{\text{rand}}\). This gives more samples in narrow passages. This
technique could easily be applied to on-line RRT planners.
\begin{algorithm}[ht!]
\caption{Retraction-based RRT Extension}
\label{alg:rrrt}
\begin{algorithmic}[1]
    \STATE \(q_r \leftarrow \text{a random configuration in \(C_{\text{space}}\)}\)
    \STATE \(q_n \leftarrow \text{the nearest neighbor of \(q_r\) in \(T\)}\)
    \IF{\(\operatorname{CollisionFree}(q_n,q_r)\)}
        \STATE \(T.\operatorname{addVertex}(q_r)\)
        \STATE \(T.\operatorname{addEdge}(q_n,q_r)\)
    \ELSE
        \STATE \(S \leftarrow \operatorname{Optimize}(q_r,q_n)\)
        \FORALL{\(q_i \in S\)}
            \STATE Standard RRT Extension\((T,q_i)\)
        \ENDFOR
    \ENDIF
    \RETURN $T$
\end{algorithmic}
\end{algorithm}

\section{Execution Extended RRT}
\label{sec:ERRT}
The Execution Extended RRT presented in~\cite{Bruce02} introduces two
extensions to RRT to build an on-line planner, the waypoint cache and adaptive
cost penalty search, which
improve re-planning efficiency and the quality of generated paths. ERRT uses
a \mbox{kd-tree}~(see section~\ref{sec:kd-tree}) to speed nearest neighbor look-up, and does not use
bidirectional search. The waypoint cache is implemented by keeping a constant
size array of states, and whenever a plan is found, all the states in the plan
are placed in the cache with random replacement. Then, when the tree is no
longer valid, a new tree must be grown, and there are three
possibilities for choosing a new target state, as shown in algorithm
\ref{alg:choosetarget}, which is used instead of
\(\operatorname{RandomConfig}()\) in previous
algorithms. With probability
P[\textit{goal}], the goal is chosen as the target; with probability
P[\textit{waypoint}], a random waypoint is chosen, and with the remaining
probability a uniform state is chosen as before. In~\cite{Bruce02} the values used
are P[\textit{goal}]$=0.1$ and P[\textit{waypoint}]$=0.6$.

Another extension is adaptive cost penalty search, where the planner adaptively
modified a parameter to help it find shorter paths. A value of 1 for beta will
always extend from the root node, while a value of 0 is equivalent to the
original algorithm. However, the paper~\cite{Bruce02} lacks
implementation details and experimental results on this extension.

\begin{algorithm}[ht!]
    \caption{$\operatorname{ChooseTarget}(q,{\text{goal}})$}
    \label{alg:choosetarget}
    \begin{algorithmic}[1]
        \STATE \(p \leftarrow \operatorname{UniformRandom}(0.0,1.0)\)
        \STATE \(i \leftarrow \operatorname{UniformRandom}(0.0,\text{NumWayPoints})\)
        \IF{\(0 < p < \text{GoalProb}\)}
            \RETURN \(q_{\text{goal}}\)
        \ELSIF{\(\text{GoalProb}<p<\text{GoalProb}+\text{WayPointProb}\)}
            \RETURN \(\text{WayPointCache}[i]\)
        \ELSIF{\(\text{GoalProb}+\text{WayPointProb}<p<1\)}
            \RETURN \(\text{RandomConfig}()\)
        \ENDIF
    \end{algorithmic}
\end{algorithm}

\section{Dynamic RRT}
\label{sec:DRRT}
The Dynamic Rapidly-Exploring Random Tree described in~\cite{Ferguson06} is a
probabilistic analog to the widely used D* family of algorithms. It works by
growing a tree from \(q_{\text{goal}}\) to \(q_{\text{init}}\), as shown in algorithm
\ref{alg:drrt}. This has the advantage that the
root of the tree does not have to be moved during the lifetime of the planning
and execution. In some problem classes the robot has limited range
sensors, thus moving or newly appearing obstacles will be near the robot, not
near the goal. In general this strategy attempts to trim smaller branches that are
farther away from the root. When new information concerning the configuration space is
received, the algorithm removes the newly-invalid branches of the
tree (algorithms~\ref{alg:trimrrt} and~\ref{alg:invalidatenodes}), and
grows the remaining tree, focusing, with a certain probability (empirically tuned
to $0.4$ in~\cite{Ferguson06}) to a vicinity of the recently trimmed branches,
by using the waypoint
cache of the ERRT~(algorithm~\ref{alg:choosetarget}). In experiments presented
in~\cite{Ferguson06} DRRT vastly outperforms ERRT.

\begin{algorithm}[ht!]
    \caption{$\operatorname{DRRT}()$}
    \label{alg:drrt}
    \begin{algorithmic}[1]
        \STATE \(q_{\text{robot}} \leftarrow \text{the current robot position}\)
        \STATE \(T \leftarrow \operatorname{BuildRRT}(q_{\text{goal}},q_{\text{robot}})\)
        \WHILE{\(q_{\text{robot}} \neq q_{\text{goal}}\)}
            \STATE \(q_{\text{next}}  \leftarrow \operatorname{Parent}(q_{\text{robot}})\)
            \STATE Move from \(q_{\text{robot}}\) to \(q_{\text{next}}\)
            \FORALL{obstacles that changed $O$}
                \STATE \(\operatorname{InvalidateNodes}(O)\)
            \ENDFOR
            \IF{Solution path contains an invalid node}
                \STATE \(\operatorname{ReGrowRRT}()\)
            \ENDIF
        \ENDWHILE
    \end{algorithmic}
\end{algorithm}

\begin{algorithm}[ht!]
\caption{$\operatorname{ReGrowRRT}()$}
\label{alg:regrowrrt}
\begin{algorithmic}[1]
    \STATE \(\operatorname{TrimRRT}()\)
    \STATE \(\operatorname{GrowRRT}()\)
\end{algorithmic}
\end{algorithm}

\begin{algorithm}[ht!]
\caption{$\operatorname{TrimRRT}()$}
\label{alg:trimrrt}
\begin{algorithmic}[1]
    \STATE \(S \leftarrow \emptyset, i \leftarrow 1\)
    \WHILE{\(i < T.\operatorname{size}()\)}
        \STATE \(q_i \leftarrow  T.\operatorname{node}(i)\)
        \STATE \(q_p \leftarrow  \operatorname{Parent}(q_i)\)
        \IF{\(q_p.\text{flag} = \text{INVALID}\)}
            \STATE \(q_i.\text{flag} \leftarrow  \text{INVALID}\)
        \ENDIF
        \IF{\(q_i.\text{flag} \neq \text{INVALID}\)}
            \STATE \(S \leftarrow  S \bigcup \{q_i\}\)
        \ENDIF
        \STATE \(i \leftarrow i+1\)
    \ENDWHILE
    \STATE \(T \leftarrow  \operatorname{CreateTreeFromNodes}(S)\)
\end{algorithmic}
\end{algorithm}

\begin{algorithm}[ht!]
\caption{$\operatorname{InvalidateNodes}(obstacle)$}
\label{alg:invalidatenodes}
\begin{algorithmic}[1]
    \STATE \(E \leftarrow \operatorname{FindAffectedEdges}(\text{obstacle})\)
    \FORALL{\(e \in E\)}
        \STATE \(q_e \leftarrow  \operatorname{ChildEndpointNode}(e)\)
        \STATE \(q_e.\text{flag} \leftarrow  \text{INVALID}\)
    \ENDFOR
\end{algorithmic}
\end{algorithm}

\section{Multipartite RRT}
\label{sec:MPRRT}
Multipartite RRT presented in~\cite{Zucker07} is another RRT variant which
supports planning in unknown or dynamic environments. MP-RRT maintains a
forest $F$ of disconnected sub-trees which lie in \(C_{\text{free}}\), but which are not
connected to the root node \(q_{\text{root}}\) of $T$, the main tree. At the start of a
given planning iteration, any nodes of $T$ and $F$ which are no longer valid are
deleted, and any disconnected sub-trees which are created as a result are placed
into $F$~(as seen in algorithms~\ref{alg:mprrtsearch} and~\ref{alg:pruneandprepend}).
With given probabilities,
the algorithm tries to connect $T$ to a
new random state, to the goal state, or to the root of a tree in $F$ (algorithm~%
\ref{alg:selectsample}). In~%
\cite{Zucker07}, a simple greedy smoothing heuristic is used, that tries to
shorten paths by skipping intermediate nodes. The MP-RRT
is compared to an iterated RRT, ERRT and DRRT, in 2D, 3D and 4D problems, with
and without smoothing. For most of the experiments, MP-RRT modestly outperforms
the other algorithms, but in the 4D case with smoothing, the performance gap in
favor of MP-RRT is much larger. The authors explained this fact due to MP-RRT
being able to construct much more robust plans in the face of dynamic obstacle
motion. Another algorithm that utilizes the concept of forests is
Reconfigurable Random Forests (RRF) presented in~\cite{Li02}, but without the
success of MP-RRT.

\begin{algorithm}[ht!]
\caption{$\operatorname{MPRRTSearch}(q_{\text{init}})$}
\label{alg:mprrtsearch}
\begin{algorithmic}[1]
    \STATE \(T \leftarrow \text{the previous search tree, if any}\)
    \STATE \(F \leftarrow \text{the previous forest of disconnected sub-trees}\)
    \STATE \(q_{\text{init}} \leftarrow \text{the initial state}\)
    \IF{\(T=\emptyset\)}
        \STATE \(q_{\text{root}} \leftarrow  q_{\text{init}}\)
        \STATE \(\operatorname{Insert}(q_{\text{root}},T)\)
    \ELSE
        \STATE \(\operatorname{PruneAndPrepend}(T,F,q_{\text{init}})\)
        \IF{\(\operatorname{TreeHasGoal}(T)\)}
            \RETURN \TRUE
        \ENDIF
    \ENDIF
    \WHILE{search time/space remaining}
        \STATE \(q_{\text{new}} \leftarrow  \operatorname{SelectSample}(F)\)
        \STATE \(q_{\text{near}} \leftarrow  \operatorname{NearestNeighbor}(q_{\text{new},T})\)
        \IF{\(q_{\text{new}}\in F\)}
            \STATE \(b_{\text{connect}} \leftarrow  \operatorname{Connect}(q_{\text{near}},q_{\text{new}})\)
            \IF{\(b_{\text{connect}}\) \textbf{and}  \(\operatorname{TreeHasGoal}(T)\)}
                \RETURN \TRUE
            \ENDIF
        \ELSE
            \STATE \(b_{\text{extend}} \leftarrow   \operatorname{Extend}(q_{\text{near}},q_{\text{new}})\)
            \IF{\(b_{\text{extend}}\) \textbf{and} \(\operatorname{IsGoal}(q_{\text{new}})\)}
                \RETURN \TRUE
            \ENDIF
        \ENDIF
    \ENDWHILE
    \RETURN \FALSE
\end{algorithmic}
\end{algorithm}

\begin{algorithm}[ht!]
\caption{$\operatorname{PruneAndPrepend}(T,F,q_{\text{init}})$}
\label{alg:pruneandprepend}
\begin{algorithmic}[1]
    \FORALL{\(q\in T,F\)}
        \IF{\textbf{not} \(\operatorname{NodeValid}(q)\)}
            \STATE \(\operatorname{KillNode}(q)\)
        \ELSIF{\textbf{not} \(\operatorname{ActionValid}(q)\)}
            \STATE \(\operatorname{SplitEdge}(q)\)
        \ENDIF
    \ENDFOR
    \IF{\textbf{not} \(T=\emptyset\) \textbf{and} \(q_{\text{root}} \neq q_{\text{init}}\)}
        \IF{\textbf{not} \(\operatorname{ReRoot}(T,q_{\text{init}})\)}
            \STATE \(F \leftarrow  F \bigcup T\)
            \STATE \(T.\operatorname{init}(q_{\text{init}})\)
        \ENDIF
    \ENDIF
\end{algorithmic}
\end{algorithm}

\begin{algorithm}[ht!]
\caption{$\operatorname{SelectSample}(F)$}
\label{alg:selectsample}
\begin{algorithmic}[1]
    \STATE \(p \leftarrow  \operatorname{Random}(0,1)\)
    \IF{\(p<p_{\text{goal}}\)}
        \STATE \(q_{\text{new}} \leftarrow q_{\text{goal}}\)
    \ELSIF{\(p<(p_{\text{goal}}+p_{\text{forest}})\) \textbf{and not} \(\operatorname{Empty}(F)\)}
        \STATE \(q_{\text{new}} \leftarrow  q \in \operatorname{SubTreeRoots}(F)\)
    \ELSE
        \STATE \(q_{\text{new}} \leftarrow  \operatorname{RandomState}()\)
    \ENDIF
    \RETURN \(q_{\text{new}}\)
\end{algorithmic}
\end{algorithm}

\section{Rapidly Exploring Evolutionary Tree}
\label{sec:RET}
The Rapidly Exploring Evolutionary Tree, introduced in~\cite{Martin07} uses a
bidirectional RRT and a \mbox{kd-tree}~(see section~\ref{sec:kd-tree}) for efficient
nearest neighbor search. The
modifications to the \(\operatorname{Extend}()\) function are shown in algorithm
\ref{alg:extendtotarget}. The
re-balancing of a \mbox{kd-tree} is costly, and in this paper a simple threshold on the
number of nodes added before re-balancing was used. The authors suggest using the
method described in~\cite{Atramentov02} and used in~\cite{Bruce02} to improve
the search speed. The novelty in this algorithm comes from the introduction of
an evolutionary algorithm~\cite{Back97} that builds a population of biases for the RRTs. The
genotype of the evolutionary algorithm consists of a single robot configuration for each tree. This
configuration is sampled instead of the uniform distribution. To balance
exploration and exploitation, the evolutionary algorithm was designed with 50\% elitism. The fitness
function is related to the number of left and right branches traversed during
the insertion of a new node in the \mbox{kd-tree}. The goal is to introduce a bias to
the RRT algorithm which shows preference to nodes created away from the center
of the tree. The authors suggest combining RET with DRRT or MP-RRT.

\begin{algorithm}[ht!]
\caption{$\operatorname{ExtendToTarget}(T)$}
\label{alg:extendtotarget}
\begin{algorithmic}[1]
    \STATE \textbf{static} $p$: population, \(inc \leftarrow 1\)
    \STATE \(p'\): temporary population
    \IF{\(\text{inc}>\operatorname{length}(p)\)}
        \STATE \(\operatorname{SortByFitness}(p)\)
        \STATE \(p' \leftarrow  \text{null}\)
        \FORALL{\(i \in p\)}
            \IF{i is in upper 50\%}
                \STATE \(\operatorname{AddIndividual}(i,p')\)
            \ELSE
                \STATE \(i \leftarrow  \operatorname{RandomState}()\)
                \STATE \(\operatorname{AddIndividual}(i,p')\)
            \ENDIF
        \ENDFOR
        \STATE \(p \leftarrow  p'\)
        \STATE \(\text{inc} \leftarrow  1\)
    \ENDIF
    \STATE \(q_r \leftarrow  p(\text{inc})\)
    \STATE \(q_{\text{near}} \leftarrow   \operatorname{Nearest}(T,q_r)\)
    \STATE \(q_{\text{new}} \leftarrow  \operatorname{Extend}(T,q_{\text{near}})\)
    \IF{\(q_{\text{new}} \neq \emptyset\)}
        \STATE \(\operatorname{AddNode}(T,q_{\text{new}})\)
        \STATE \(\operatorname{AssignFitness}(p(\text{inc}),\operatorname{fitness}(q_{\text{new}})\)
    \ELSE
        \STATE \(\operatorname{AssignFitness}(p(\text{inc}),0)\)
    \ENDIF
    \RETURN \(q_{\text{new}}\)
\end{algorithmic}
\end{algorithm}

\section{Multidimensional Binary Search Trees}
\label{sec:kd-tree}
The kd-tree, first introduced in~\cite{Bentley75}, is a binary tree in which every 
node is a \mbox{k-dimensional} point. Every
non-leaf node generates a splitting hyperplane that divides the space into two
subspaces. In the RRT algorithm, the number of points grows incrementally,
unbalancing the tree, thus slowing nearest-neighbor queries. Re-balancing a kd-tree
is costly, so in~\cite{Atramentov02} the authors present another approach:
A vector of trees is constructed, where for $n$ points there is a tree that
contains $2^i$ points for each $"1"$ in the $i^{th}$ place of the binary
representation of $n$. As bits are cleared in the representation due to
increasing $n$, the trees are deleted, and the points are included in a tree
that corresponds to the higher-order bit which is changed to $"1"$. This general
scheme incurs in logarithmic-time overhead, regardless of dimension. Experiments
show a substantial performance increase compared to a naive brute-force approach.

\section{Evolutionary Planner/Navigator}
\label{sec:EP/N}
An evolutionary algorithm~\cite{Back97} is a generic population-based meta-heuristic
optimization algorithm. It is inspired in biological evolution, using methods
such as individual selection, reproduction and mutation. The population is
composed of candidate solutions and they are evaluated according to a fitness
function.

The Evolutionary Planner/Navigator presented in~\cite{Xiao96},~\cite{Xiao97}, and~
\cite{Trojanowski97} is an evolutionary
algorithm for path finding in dynamic environments. A high level description is
shown in algorithm~\ref{alg:epn}. A difference with RRT is
that it can optimize the path according to any fitness function
defined (length, smoothness, etc), without the need for a post-processing step.
Experimental tests have shown it has
good performance for sparse maps, but no so much for difficult maps with narrow
passages or too crowded with obstacles. However, when a feasible path is found,
it is very efficient at optimizing it and adapting to the dynamic obstacles.
\begin{algorithm}[ht!]
\caption{EP/N}
\label{alg:epn}
\begin{algorithmic}[1]
    \STATE \(P(t)\): population at generation \(t\)
    \STATE \(t \leftarrow 0\)
    \STATE \(\operatorname{Initialize}(P(t))\)
    \STATE \(\operatorname{Evaluate}(P(t))\)
    \WHILE{(\textbf{not} termination-condition)}
        \STATE \(t \leftarrow t + 1\)
        \STATE Select operator $o_j$ with probability $p_j$
        \STATE Select parent(s) from $P(t)$
        \STATE Produce offspring applying $o_j$ to selected parent(s)
        \STATE Evaluate offspring
        \STATE Replace worst individual in $P(t)$ by new offspring
        \STATE Select best individual $p$ from $P(t)$
        \IF{\(\operatorname{Feasible}(p)\)}
            \STATE Move along path \(p\)
            \STATE Update all individuals in \(P(t)\) with current position
            \IF{changes in environment}
                \STATE Update map
            \ENDIF
            \STATE \(\operatorname{Evaluate}(P(t))\)
        \ENDIF
        \STATE \(t \leftarrow t + 1\)
    \ENDWHILE
\end{algorithmic}
\end{algorithm}
Every individual in the population is a sequence of nodes, representing
nodes in a path consisting of straight-line segments. Each node consists of an
$(x,y)$ pair and a state variable $b$ with information about the feasibility of
the point and the path segment connecting it to the next point. Individuals have
variable length.

Since a path $p$ can be either feasible or unfeasible, two evaluation functions
are used. For feasible paths (equation~\ref{eq:fitness}), the goal is to minimize distance traveled,
maintain a smooth trajectory and satisfy a clearance requirement (the robot
should not approach the obstacles too closely). For unfeasible paths, we use
equation~\ref{eq:unfeasible}, taken from~\cite{Xiao97-2}, where $\mu$ is the number 
of intersections of a whole path with obstacles and $\eta$ is the average number
of intersections per unfeasible segment.

\begin{equation}
\label{eq:fitness}
\operatorname{eval}_f(p)=w_d\cdot \operatorname{dist}(p) + w_s\cdot \operatorname{smooth}(p)+w_c\cdot \operatorname{clear}(p)
\end{equation}

\begin{equation}
\label{eq:unfeasible}
\operatorname{eval}_u(p)= \mu + \eta
\end{equation}

\begin{figure}[ht!]
\begin{center}
\includegraphics[width=0.9\textwidth]{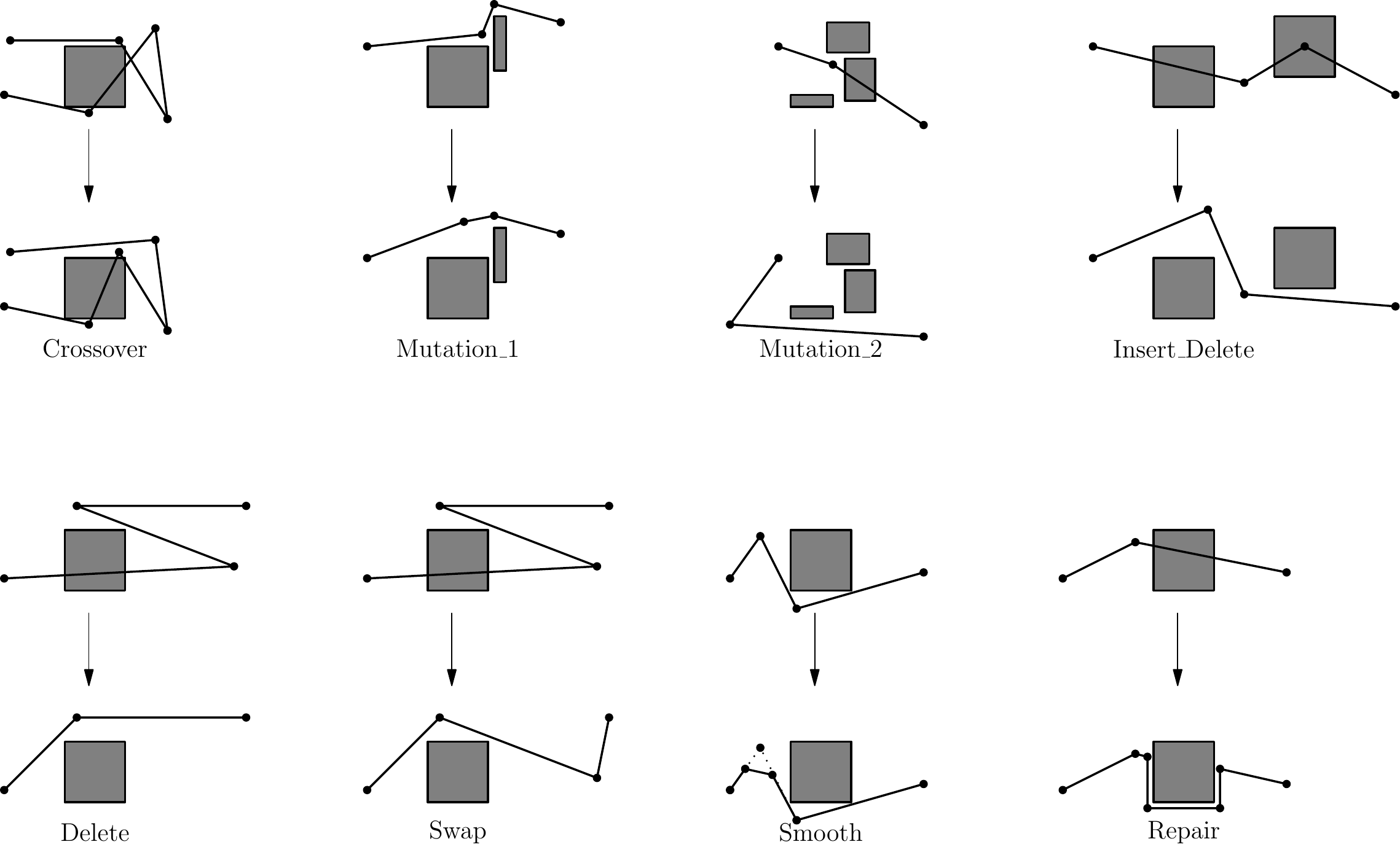}
\caption{The roles of the genetic operators}
\label{fig:operators}
\end{center}
\end{figure}
EP/N uses eight different operators, as shown in figure~%
\ref{fig:operators} (description taken from~\cite{Xiao96}):
\begin{description}
\item{\it Crossover:} Recombines two (parent) paths into two new paths. The
parent paths are divided randomly into two parts respectively and recombined:
The first part of the first path with the second part of the second path, and
the first part of the second path with the second part of the first path. Note
that there can be different numbers of nodes in the two parent paths.
\item{\it Mutate\_1:} Used to fine tune node coordinates in a
feasible path for shape adjustment. This operator randomly adjusts node
coordinates within some local clearance of the path so that the path remains
feasible afterwards.
\item{\it Mutate\_2:} Used for large random changes of node coordinates in
a path, which can be either feasible or unfeasible.
\item{\it Insert-Delete:} Operates on an unfeasible path by inserting
randomly generated new nodes into unfeasible path segments and deleting
unfeasible nodes (i.e., path nodes that are inside obstacles).
\item{\it Delete:} Deletes nodes from a path, which can be
either feasible or unfeasible. If the path is unfeasible,
the deletion is done randomly. Otherwise, the operator
decides whether a node should definitely be deleted
based on some heuristic knowledge, and if a node is
not definitely deletable, its deletion will be random.
\item{\it Swap:} Swaps the coordinates of randomly selected
adjacent nodes in a path, which can be either feasible
or unfeasible.
\item{\it Smooth:} Smoothens turns of a feasible path by ``cutting corners,''
i.e., for a selected node, the operator
inserts two new nodes on the two path segments connected to that node
respectively and deletes that selected node. The nodes with sharper turns are
more likely to be selected.
\item{\it Repair:} Repairs a randomly selected unfeasible segment
in a path by ``pulling'' the segment around its
intersecting obstacle.
\end{description}
The probabilities of using each operator is set randomly at the beginning, and
then are updated according to the success ratio of each operator, so more
successful operators are used more often, and automatically chosen according to
the instance of the problem, eliminating the difficult problem of hand tuning
the probabilities.

In~\cite{Trojanowski97}, the authors include a memory buffer for each
individual to store good paths from its ancestors, which gave a small performance
gain.

In~\cite{Elshamli04}, the authors propose strategies for improving the
stability
and controlling population diversity for a simplified version of the EP/N.
An improvement proposed by the authors in~\cite{Xiao97} is using heuristics for the initial
population, instead of random initialization. We will consider this improvement
in section~\ref{sec:RRT-EP/N}.

Other evolutionary algorithms have also been proposed for similar problems, in~%
\cite{Nagib04} a binary genetic algorithm is used for an offline planner, and~%
\cite{Nikolos03} presents an algorithm to generate curved trajectories in 3D
space for an unmanned aerial vehicle.

EP/N has been adapted to an 8-connected grid model in~%
\cite{Alfaro08} (with previous work in~\cite{Alfaro05} and~\cite{Alfaro05-2}).
The authors study two different crossover operators and four asexual operators.
Experimental results for this new algorithm (EvP) in static unknown environments
show that it is faster than EP/N.

\chapter{Proposed Techniques}

\section{Combining RRT and EP/N}
\label{sec:RRT-EP/N}
As mentioned in section~\ref{sec:stateoftheart}, RRT variants produce
suboptimal solutions, which must later be post-processed for shortening,
smoothing or other desired characteristics. On the other hand, EP/N, 
presented in section~\ref{sec:EP/N}, can optimize a solution according to any
given fitness function. However, this algorithm is slower at finding a first 
feasible solution. In this section we propose a combined approach, that uses
RRT to find an initial solution to be used as starting point for EP/N,
taking advantage of the strong points of both algorithms.

\subsection{The Combined Strategy}

\subsubsection{Initial Solution}
EP/N as presented in section~\ref{sec:EP/N} can not find feasible paths in a
reasonable amount of time in any but very sparse maps. For this reason, RRT
will be used to generate a first initial solution, ignoring the effects produced
by dynamic objects. This solution will be in the initial population of the
evolutionary algorithm, along with random solutions.

\subsubsection{Feasibility and Optimization}
EP/N is the responsible of regaining feasibility when it is lost due to a
moving obstacle or a new obstacle found in a partially known or totally unknown
environment. If a feasible solution can not be found in a given amount of time,
the algorithm is restarted, keeping its old population, but adding a new
individual
generated by RRT.

\subsection{Algorithm Implementation}
\begin{algorithm}[ht]
    \caption{$\operatorname{Main}()$}
    \label{alg:rrtepn}
    \begin{algorithmic}[1]
        \STATE \(q_{\text{robot}} \leftarrow \text{is the current robot position}\)
        \STATE \(q_{\text{goal}} \leftarrow \text{is the goal position}\)
        \WHILE{\(q_{\text{robot}} \neq q_{\text{goal}}\)}
             \STATE \(\operatorname{updateWorld}(\text{time})\)
             \STATE \(\operatorname{processRRTEPN}(\text{time})\)
        \ENDWHILE
    \end{algorithmic}
\end{algorithm}

The combined RRT-EP/N algorithm proposed here works by alternating environment
updates and path planning, as can be seen in
algorithm~\ref{alg:rrtepn}. The first stage of the path planning (see algorithm~%
\ref{alg:processrrtepn})
is to find an initial path using a RRT technique, ignoring any cuts that might happen during
environment updates. Thus, the RRT ensures that the path found
does not collide with static obstacles, but might collide with dynamic obstacles in the future.
When a first path is found,
the navigation is done by using the standard EP/N as shown in algorithm~%
\ref{alg:epn}.

\begin{algorithm}[ht]
    \caption{$\operatorname{processRRTEPN}(time)$}
    \label{alg:processrrtepn}
    \begin{algorithmic}[1]
        \STATE \(q_{\text{robot}} \leftarrow \text{the current robot position}\)
        \STATE \(q_{\text{start}} \leftarrow \text{the starting position}\)
        \STATE \(q_{\text{goal}} \leftarrow \text{the goal position}\)
        \STATE \(T_{\text{init}} \leftarrow \text{the tree rooted at the robot position}\)
        \STATE \(T_{\text{goal}} \leftarrow \text{the tree rooted at the goal position}\)
        \STATE \(\text{path} \leftarrow \text{the path extracted from the merged RRTs}\)
        \STATE \(q_{\text{robot}} \leftarrow q_{\text{start}}\)
        \STATE \(T_{\text{init}}.\operatorname{init}(q_{\text{robot}})\)
        \STATE \(T_{\text{goal}}.\operatorname{init}(q_{\text{goal}})\)
        \WHILE{time elapsed \(<\) time}
            \IF{First path not found}
                \STATE \(\operatorname{RRT}(T_{\text{init}},T_{\text{goal}})\)
            \ELSE
                \STATE \(\operatorname{EP/N}()\)
            \ENDIF
        \ENDWHILE
    \end{algorithmic}
\end{algorithm}

\section{A Simple Multi-stage Probabilistic Algorithm}
\label{sec:RRT-LP}

In highly dynamic environments, with many (or a few but fast) relatively small
moving obstacles, regrowing trees are pruned too fast, cutting away important
parts of the trees before they can be replaced. This dramatically reduces
the performance of the algorithms, making them unsuitable for these classes
of problems.
We believe that better performance could be obtained
by slightly modifying a RRT solution using simple obstacle-avoidance
operations on the new colliding points of the path by informed local search.
The path could be greedily optimized if the path has reached the feasibility
condition.

\subsection{A Multi-stage Probabilistic Strategy}

If solving equation~\ref{eq:problem} is not a simple task in static environments,
solving dynamic versions turns out to be even more difficult. In dynamic path planning
we cannot wait until reaching the optimal solution because we must deliver
a ``good enough'' plan within some time restriction. Thus, a heuristic approach
must be developed to tackle the on-line nature of the problem. The heuristic
algorithms presented in sections~\ref{sec:ERRT}, \ref{sec:DRRT} and~\ref{sec:MPRRT}
extend a method developed for static environments, which produces poor response
to highly dynamic environments and unwanted complexity of the algorithms.

We propose a multi-stage combination of simple heuristic and probabilistic techniques
to solve each part of the problem: Feasibility, initial solution and optimization.

\begin{figure}[ht]
\begin{center}
\includegraphics[width=0.6\textwidth]{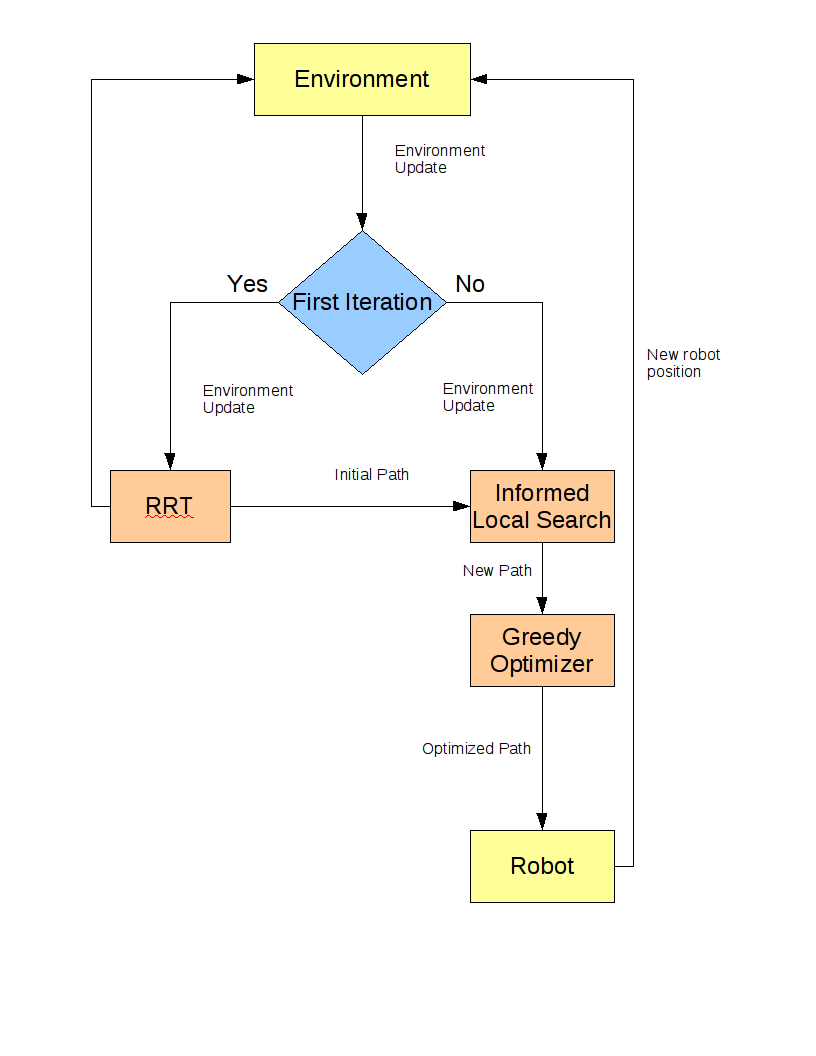}
\caption[A Multi-stage Strategy for Dynamic Path Planning]{\textbf{A Multi-stage Strategy for Dynamic Path Planning}. This figure describes
the life-cycle of the multi-stage algorithm presented here. The RRT, informed local search, and greedy heuristic are combined
to produce a cheap solution to the dynamic path planning problem.}
\label{fig:diag}
\end{center}
\end{figure}

\subsubsection{Feasibility}
The key point in this problem is the hard constraint in equation~\ref{eq:problem}
which must be met before even thinking about optimizing. The problem is that in
highly dynamic environments a path turns rapidly from feasible to unfeasible
--- and the other way around --- even if our path does not change.
We propose a simple \emph{informed local search} to obtain paths in
\(C_{\text{free}}\).
The idea is to randomly search for a \(C_{\text{free}}\) path by modifying the nearest colliding
segment of the path. As we include in the search some knowledge of the problem,
the \emph{informed} term is coined to distinguish it from blind local search.
The details of the operators used for the modification of the path are described in
section~\ref{sec:implementationrrtlp}. If a feasible solution can not be found in a
given amount of time, the algorithm is restarted, with a new starting point generated by a RRT variant.
\subsubsection{Initial Solution}
The problem with local search algorithms is that they repair a solution that it is
assumed to be near the feasibility condition.
Trying to produce feasible paths from scratch with local search (or even
with evolutionary algorithms~\cite{Xiao97}) is not a good idea due the randomness of the
initial solution. Therefore, we propose feeding the informed local search with a \emph{standard RRT} solution
at the start of the planning, as can be seen in figure~\ref{fig:diag}.
\subsubsection{Optimization}
Without an optimization criterion, the path could grow infinitely large in time or
size. Therefore, the \(\operatorname{eval}(\cdot,\cdot)\) function must be minimized when a
(temporary) feasible path is obtained. A simple greedy technique is used
here: We test each point in the solution to check if it can be removed
maintaining feasibility; if so, we remove it and check the following point,
continuing until reaching the last one.

\subsection{Algorithm Implementation}
\label{sec:implementationrrtlp}

\begin{algorithm}[ht]
    \caption{$\operatorname{Main}()$}
    \label{alg:rrtlp}
    \begin{algorithmic}[1]
        \STATE \(q_{\text{robot}} \leftarrow \text{the current robot position}\)
        \STATE \(q_{\text{goal}} \leftarrow \text{the goal position}\)
        \WHILE{\(q_{\text{robot}} \neq q_{\text{goal}}\)}
             \STATE \(\operatorname{updateWorld}(\text{time})\)
             \STATE \(\operatorname{processMultiStage}(\text{time})\)
        \ENDWHILE
    \end{algorithmic}
\end{algorithm}

The multi-stage algorithm proposed in this thesis works by alternating environment
updates and path planning, as can be seen in
algorithm~\ref{alg:rrtlp}. The first stage of the path planning~(see algorithm~%
\ref{alg:processmultistage})
is to find an initial path using a RRT technique, ignoring any cuts that might happen during
environment updates. Thus, RRT ensures that the path found
does not collide with static obstacles, but might collide with dynamic obstacles in the future.
When a first path is found,
the navigation is done by alternating a simple informed local search and
a simple greedy heuristic as shown in figure~\ref{fig:diag}.

\begin{algorithm}[ht]
    \caption{$\operatorname{processMultiStage}(\text{time})$}
    \label{alg:processmultistage}
    \begin{algorithmic}[1]
        \STATE $q_{\text{robot}} \leftarrow$ is the current robot position
        \STATE $q_{\text{start}} \leftarrow$ is the starting position
        \STATE $q_{\text{goal}} \leftarrow$ is the goal position
        \STATE $T_{\text{init}} \leftarrow$ is the tree rooted at the robot position
        \STATE $T_{\text{goal}} \leftarrow$ is the tree rooted at the goal position
        \STATE $\text{path} \leftarrow$ is the path extracted from the merged RRTs
        \STATE $q_{\text{robot}} \leftarrow q_{\text{start}}$
        \STATE $T_{\text{init}}.\operatorname{init}(q_{\text{robot}})$
        \STATE $T_{\text{goal}}.\operatorname{init}(q_{\text{goal}})$
        \WHILE{time elapsed $<$ time}
            \IF{First path not found}
                \STATE \(\operatorname{RRT}(T_{\text{init}},T_{\text{goal}})\)
            \ELSE
                \IF{path is not collision free}
                    \STATE firstCol $\leftarrow$ collision point closest to robot
                    \STATE \(\operatorname{arc}(\text{path}, \text{firstCol})\)
                    \STATE \(\operatorname{mut}(\text{path}, \text{firstCol})\)
                \ENDIF
            \ENDIF
        \ENDWHILE
        \STATE \(\operatorname{postProcess}(\text{path})\)
    \end{algorithmic}
\end{algorithm}

\begin{figure}[ht]
\begin{center}
\includegraphics[width=0.5\textwidth]{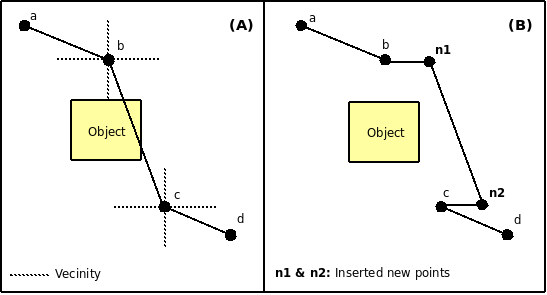}
\caption[The arc operator]{\textbf{The arc operator}. This operator draws an offset value $\Delta$ over a fixed interval called vicinity.
Then, one of the two axes is selected to perform the arc and two new consecutive points are added to the path.
$n_1$ is placed at a $\pm \Delta$ of the point $b$ and $n_2$ at $\pm \Delta$ of point $c$, both of them over the same selected axis.
The axis, sign and value of $\Delta$ are chosen randomly from an uniform distribution.}
\label{fig:arc}
\end{center}
\end{figure}

\begin{figure}[ht]
\begin{center}
\includegraphics[width=0.5\textwidth]{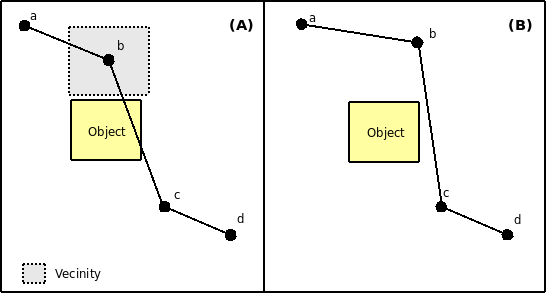}
\caption[The mutation operator]{\textbf{The mutation operator}. This operator draws two offset values $\Delta_x$ and $\Delta_y$ over a vicinity
region. Then the same point $b$ is moved in both axes from $b=[b_x,b_y]$ to $b'=[b_x \pm \Delta_x, b_y\pm \Delta_y]$, where the sign and offset values
are chosen randomly from an uniform distribution.}
\label{fig:mut}
\end{center}
\end{figure}

The second stage is the informed local search,
which is a two step function composed by the \emph{arc} and \emph{mutate} operators
(algorithms~\ref{alg:arc} and~\ref{alg:mut}).
The first one tries to build a square arc around an
obstacle, by inserting two new points between two points in the path that form a
segment colliding with an obstacle, as shown in figure~\ref{fig:arc}.
The second step in the function is a mutation
operator that moves a point close to an obstacle to a
random point in the vicinity, as explained graphically in figure~\ref{fig:mut}.
The mutation operator is inspired by the ones used in the Adaptive Evolutionary
Planner/Navigator (EP/N) presented in~\cite{Xiao97}, while the arc operator is
derived from the arc operator in the Evolutionary Algorithm presented in~%
\cite{Alfaro05}.

\begin{algorithm}[ht]
    \caption{$\operatorname{arc}(\text{path}, \text{firstCol})$}
    \label{alg:arc}
    \begin{algorithmic}[1]
        \STATE \(\text{vicinity} \leftarrow \text{some vicinity size}\)
        \STATE \(\text{randDev} \leftarrow \operatorname{random}(-\text{vicinity},\text{vicinity})\)
        \STATE \(\text{point1} \leftarrow \text{path}[\text{firstCol}]\)
        \STATE \(\text{point2} \leftarrow \text{path}[\text{firstCol}+1]\)
        \IF{\(\operatorname{random}() \% 2\)}
            \STATE \(\text{newPoint1} \leftarrow (\text{point1}[X]+\text{randDev},\text{point1}[Y])\)
            \STATE \(\text{newPoint2} \leftarrow (\text{point2}[X]+\text{randDev},\text{point2}[Y])\)
        \ELSE
            \STATE \(\text{newPoint1} \leftarrow (\text{point1}[X],\text{point1}[Y]+\text{randDev})\)
            \STATE \(\text{newPoint2} \leftarrow (\text{point2}[X],\text{point2}[Y]+\text{randDev})\)
        \ENDIF
        \IF{path segments point1-newPoint1-newPoint2-point2 are collision free}
            \STATE Add new points between point1 and point2
        \ELSE
            \STATE Drop new points
        \ENDIF
    \end{algorithmic}
\end{algorithm}

\begin{algorithm}[ht]
    \caption{$\operatorname{mut}(\text{path}, \text{firstCol})$}
    \label{alg:mut}
    \begin{algorithmic}[1]
        \STATE vicinity $\leftarrow$ some vicinity size
        \STATE path[firstCol][X] $+=$ random$(-\text{vicinity}, \text{vicinity})$
        \STATE path[firstCol][Y] $+=$ random$(-\text{vicinity}, \text{vicinity})$
        \IF{path segments before and after path[firstCol] are collision free}
            \STATE Accept new point
        \ELSE
            \STATE Reject new point
        \ENDIF
    \end{algorithmic}
\end{algorithm}

The third and last stage is the greedy optimization heuristic,
which can be seen as a post-processing for path shortening, that
eliminates intermediate nodes if doing so does not create collisions,
as is described in the algorithm~\ref{alg:postProcess}.

\begin{algorithm}[ht]
    \caption{postProcess$(path)$}
    \label{alg:postProcess}
    \begin{algorithmic}[1]
        \STATE \(i \leftarrow 0\)
        \WHILE{\(i < \operatorname{path.size}() - 2\)}
            \IF{segment \(\operatorname{path}[i]
                \text{\ to\ } \operatorname{path}[i + 2]
                \text{\ is collision free}\)}
                \STATE Delete path[i+1]
            \ELSE
                \STATE \(i \leftarrow i + 1\)
            \ENDIF
        \ENDWHILE
    \end{algorithmic}
\end{algorithm}

\chapter{Experimental Setup and Results}

\section{Experimental Setup}
\label{sec:experimentalsetup}

Although the algorithms developed in this thesis are aimed at dynamic
environments, for the sake of completeness they will also be compared in
partially known environments and in totally unknown environments, where some or
all of the obstacles become visible to the planner as the robot approaches each
one of them, simulating a robot with limited sensor range.

\subsection{Dynamic Environment}

\begin{figure}[h!]
\begin{center}
\includegraphics[width=0.7\textwidth]{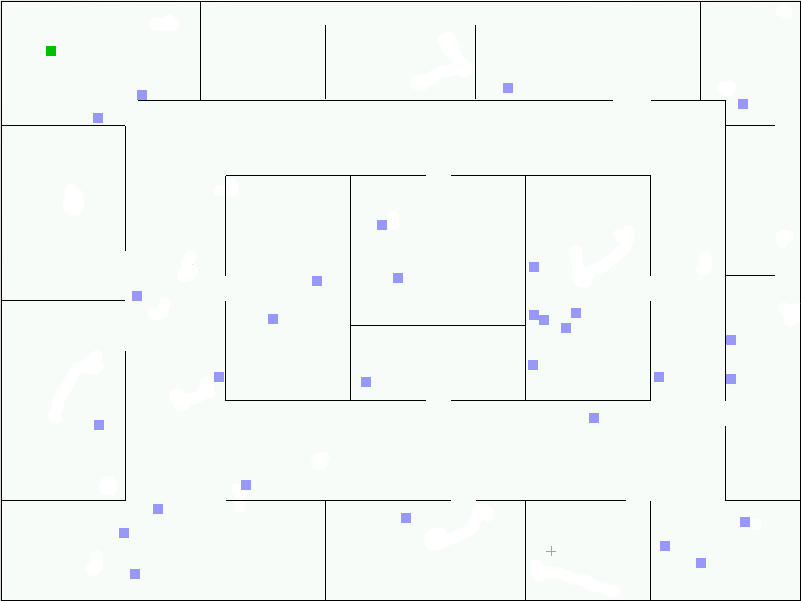}
\caption[The dynamic environment, map 1]{The dynamic environment, map 1. The \emph{green} square is our robot,
currently at the start position. The
\emph{blue} squares are the moving obstacles. The \emph{blue} cross is the goal.}
\label{fig:office-dynamic}
\end{center}
\end{figure}

\begin{figure}[h!]
\begin{center}
\includegraphics[width=0.7\textwidth]{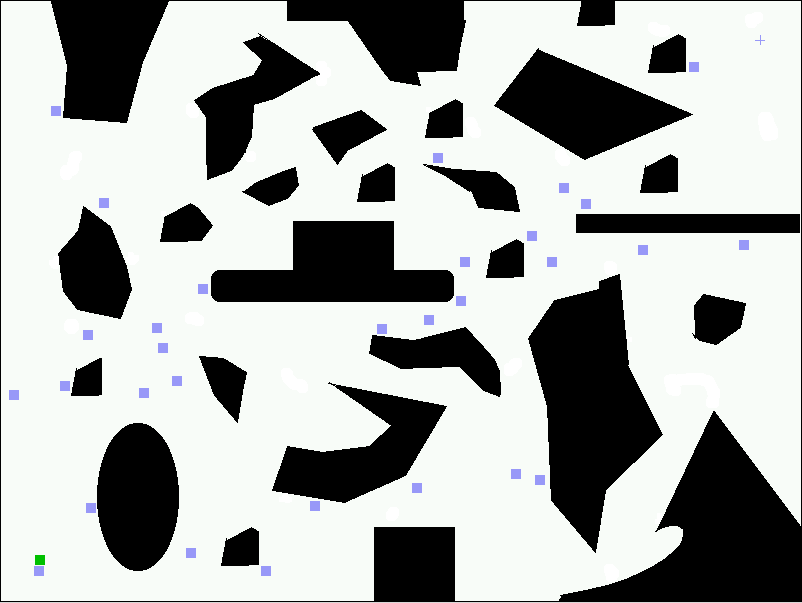}
\caption[The dynamic environment, map 2]{The dynamic environment, map 2. The \emph{green} square is our robot,
currently at the start position. The
\emph{blue} squares are the moving obstacles. The \emph{blue} cross is the goal.}
\label{fig:800-dynamic}
\end{center}
\end{figure}

The first environment for our experiments consists on two maps with 30~moving
obstacles the same size of the robot, with a random speed between 10\% and 55\%
the speed of the robot. Good performance in this environment is the main focus
of this thesis. This \emph{dynamic environments} are illustrated in figures~%
\ref{fig:office-dynamic} and~\ref{fig:800-dynamic}.

\subsection{Partially Known Environment}

\begin{figure}[h!]
\begin{center}
\includegraphics[width=0.7\textwidth]{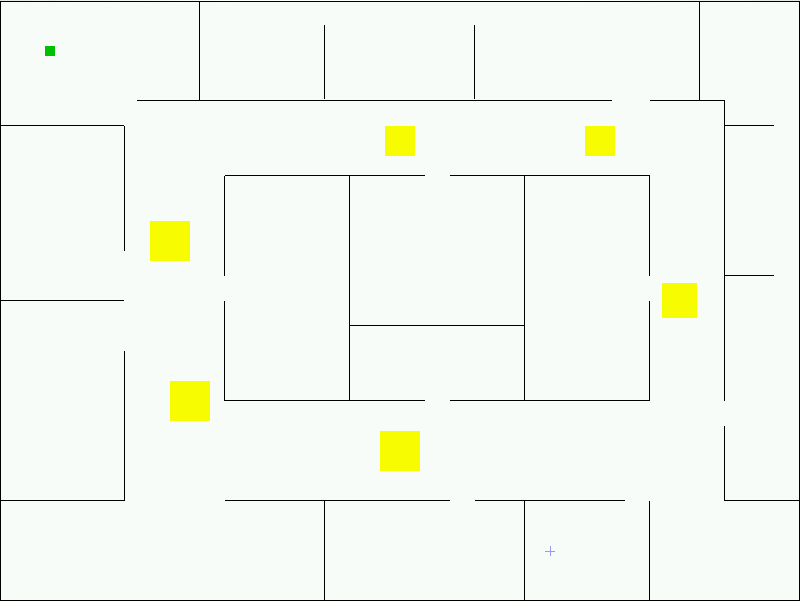}
\caption[The partially known environment, map 1]{The partially known environment, map 1. The \emph{green} square is our robot,
currently at the start position. The \emph{yellow} squares are the suddenly
appearing obstacles. The \emph{blue} cross is the goal.}
\label{fig:office-partial}
\end{center}
\end{figure}

\begin{figure}[h!]
\begin{center}
\includegraphics[width=0.7\textwidth]{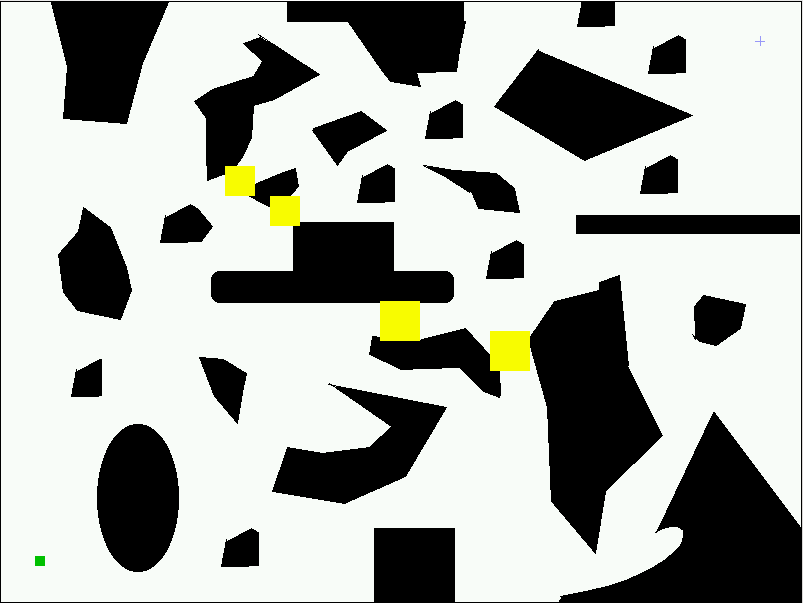}
\caption[The partially known environment, map 2]{The partially known environment, map 2. The \emph{green} square is our robot,
currently at the start position. The \emph{yellow} squares are the suddenly
appearing obstacles. The \emph{blue} cross is the goal.}
\label{fig:800-partial}
\end{center}
\end{figure}

The second environment uses the same maps, but with a few  obstacles, three to four
times the size of the robot, that become visible when the robot approaches each
one of them. This is the kind of environment that most dynamic RRT variants were designed
for. The
\emph{partially known environments} are illustrated in figure~\ref{fig:office-partial}
and~\ref{fig:800-partial}.

\subsection{Unknown Environment}

\begin{figure}[h!]
\begin{center}
\includegraphics[width=0.7\textwidth]{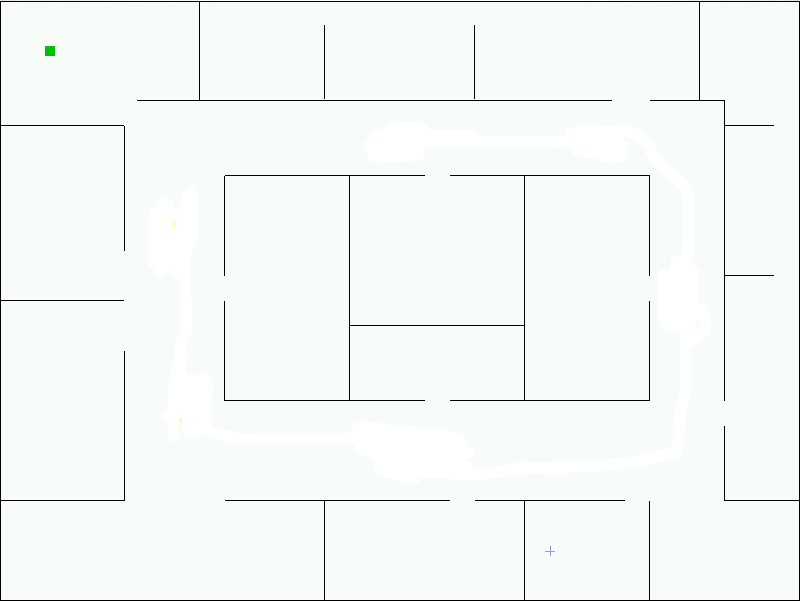}
\caption[The unknown environment]{The unknown environment. The \emph{green} square is our robot,
currently at the start position. The \emph{blue} cross is the goal. None of the
obstacles is visible initially to the planners}
\label{fig:office-unknown}
\end{center}
\end{figure}

For completeness sake, we will compare the different technique in a third
environment, were we use one of  the maps presented before, but
all the obstacles will initially be unknown to the 
planners, and will become visible as the robot approaches them, forcing several
re-plans. This \emph{unknown environment} is illustrated in figure~%
\ref{fig:office-unknown}.

\section{Implementation Details}

The algorithms where implemented in C++ using the MoPa framework%
\footnote{MoPa homepage:
\texttt{https://csrg.inf.utfsm.cl/twiki4/bin/view/CSRG/MoPa}}
 partly developed by the author. This framework features exact collision
 detection, three different map formats (including \texttt{.pbm} images from any
 graphic editor), dynamic, unknown
 and partially known environments and support for easily adding new planners.
 One of the biggest downsides is that it only supports rectangular objects, so
 several objects must be used to represent other geometrical shapes, as in
 figure~\ref{fig:800-partial}, composed of 1588 rectangular objects.

There are several variations that can be found in the literature when
implementing RRT. For all our RRT variants, the following are the details on
where we departed from the basics:
\begin{enumerate}
\item We always use two trees rooted at $q_{init}$ and $q_{goal}$.
\item Our EXTEND function, if the point cannot be added without collisions to a
tree, adds the mid point between the nearest tree node and the nearest collision
point to it.
\item In each iteration, we try to add the new randomly generated point to both
trees, and if successful in both, the trees are merged, as proposed in~%
\cite{Kuffner00}.
\item\label{sec:advance} We believe that there might be significant performance 
differences between allowing or not allowing the robot to
advance towards the node nearest to the goal when the trees are disconnected, as
proposed in~\cite{Zucker07}. 
\end{enumerate}
In point~\ref{sec:advance} above, the problem is that the robot would become stuck
if it enters a small concave zone of the environment (like a room in a building)
while there are moving obstacles inside that zone, but otherwise it can lead to
better performance. Therefore we present results for both kinds of behavior:
DRRT-adv and MP-RRT-adv move even when the trees are disconnected, while
DRRT-noadv and MP-RRT-noadv only
move when the trees are connected.

In MP-RRT, the forest was handled by simply replacing the oldest tree in it if the
forest had reached the maximum allowed size.

Concerning the parameter selection, the probability for selecting a point in the
vicinity of a point in the waypoint cache in DRRT was set to 0.4 as suggested in~%
\cite{Ferguson06}. The probability for trying to reuse a subtree in
MP-RRT was set to 0.1 as suggested in~\cite{Zucker07}.
Also, the forest size was set to 25 and the minimum size of a
tree to be saved in the forest was set to 5 nodes.

For the combined RRT-EP/N, it was considered the planner was stuck after two
seconds without a feasible solution in the population, at which point a
new solution from a RRT variant is inserted into the population. For the simple
multi-stage probabilistic algorithm, the restart is made after one second of
encountering the same obstacle along the planned path. This second approach,
which seems better, cannot be applied to the RRT-EP/N, because there is no
single path to check for collisions, but instead a population of paths. The
restart times where manually tuned.

\section{Results}
\label{sec:results}
The three algorithms were run a hundred times in each
environment and map combination.
The cutoff time was five minutes for all tests, after which the robot was
considered not to have reached the goal. Results are presented concerning:
\begin{itemize}
\item {\it Success rate (S.R.):} The percentage of times the robot arrived at the goal,
before reaching the five minutes cutoff time. This does not account for
collisions or time the robot was stopped waiting for a plan.
\item {\it Number of nearest neighbor lookups performed by each algorithm (N.N.):} One of the
possible bottlenecks for tree-based algorithms
\item {\it Number of collision checks performed (C.C.),} which in our specific
implementation takes a significant percentage of
the running time
\item {\it Time} it took the robot to reach the goal, $\pm$ the standard
deviation.
\end{itemize}

\subsection{Dynamic Environment Results}
\begin{figure}[h!]
\begin{center}
\includegraphics[width=0.9\textwidth]{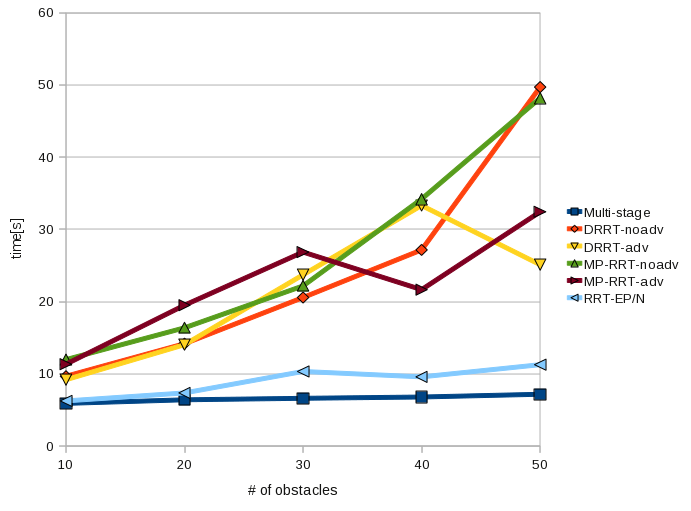}
\caption[Dynamic environment time]{Times for different number of moving obstacles in map 1.}
\label{fig:times}
\end{center}
\end{figure}

\begin{figure}[h!]
\begin{center}
\includegraphics[width=0.9\textwidth]{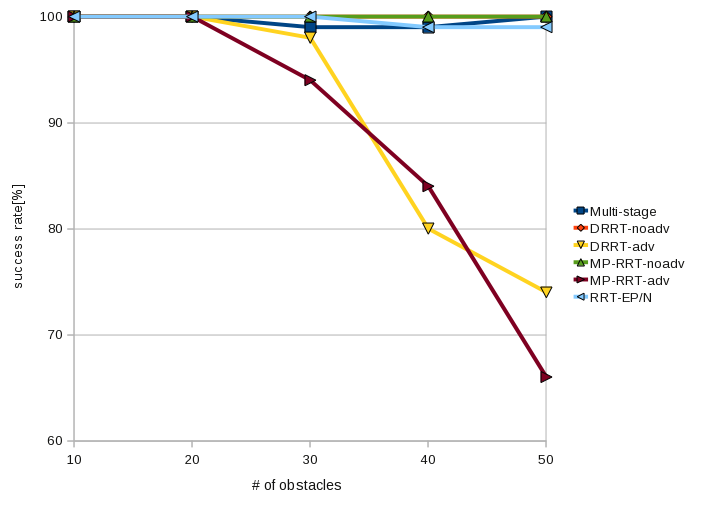}
\caption[Dynamic environment success rate]{Success rate for different number of moving obstacles in
map 1.}
\label{fig:success-rate}
\end{center}
\end{figure}

The results in tables~\ref{table:dynamic1} and~\ref{table:dynamic2} show that
the multi-stage algorithm takes considerably less time than the DRRT 
and MP-RRT to reach
the goal, with far less collision checks. The combined RRT-EP/N is a close
second. It was expected that nearest neighbor
lookups would be much lower in both combined algorithms than in the RRT variants,
because they are only performed in the initial phase and restarts, 
not during navigation.
The combined algorithms produce more consistent results
within a map,
as shown by their smaller standard deviations, but also across different
maps. An interesting fact is that in map~1 DRRT is slightly faster than MP-RRT, and
in map~2 MP-RRT is faster than DRRT. However the differences are too small to
draw any conclusions.
Figures~\ref{fig:times} and~\ref{fig:success-rate} show the times and
success rates of the different algorithms, when changing the number of dynamic
obstacles in map~1. The simple multi-stage algorithm and the mixed RRT-EP/N
clearly show the best performance, while the DRRT-adv and MP-RRT-adv
significantly reduce their success rate when confronted to more than 30~moving
obstacles.

\begin{table}[h!]
\caption{Dynamic Environment Results, map 1.}
\label{table:dynamic1}
\centering
\begin{tabular}{|l||r|r|r|r@{$\ \pm\ $}l|}
\hline
\textbf{Algorithm} & \textbf{S.R.[\%]} & \textbf{C.C.} & \textbf{N.N.} &
\multicolumn{2}{c|}{\textbf{Time[s]}}\\
\hline
Multi-stage & 99 & 23502 & 1122 & 6.62 & 0.7\\
\hline
RRT-EP/N & 100 & 58870 & 1971 & 10.34 & 14.15 \\
\hline
DRRT-noadv & 100 & 91644 & 4609 & 20.57 & 20.91\\
\hline
DRRT-adv & 98 & 107225 & 5961 & 23.72 & 34.33\\
\hline
MP-RRT-noadv & 100 & 97228 & 4563 & 22.18 & 14.71\\
\hline
MP-RRT-adv & 94 & 118799 & 6223 & 26.86 & 41.78\\
\hline
\end{tabular}
\end{table}

\begin{table}[h!]
\caption{Dynamic Environment Results, map 2.}
\label{table:dynamic2}
\centering
\begin{tabular}{|l||r|r|r|r@{$\ \pm\ $}l|}
\hline
\textbf{Algorithm} & \textbf{S.R.[\%]} & \textbf{C.C.} & \textbf{N.N.} &
\multicolumn{2}{c|}{\textbf{Time[s]}}\\
\hline
Multi-stage & 100 & 10318 & 563 & 8.05 & 1.47\\
\hline
RRT-EP/N & 100 & 21785 & 1849 & 12.69 & 5.75 \\
\hline
DRRT-noadv & 99 & 134091 & 4134 & 69.32 & 49.47\\
\hline
DRRT-adv & 100 & 34051 & 2090 & 18.94 & 17.64\\
\hline
MP-RRT-noadv & 100 & 122964 & 4811 & 67.26 & 42.45\\
\hline
MP-RRT-adv & 100 & 25837 & 2138 & 16.34 & 13.92\\
\hline
\end{tabular}
\end{table}

\subsection{Partially Known Environment Results}
Taking both maps into consideration, the results in tables~\ref{table:partial1}
and~\ref{table:partial2} 
show that both combined algorithms are faster and more consistent than the RRT
variants, with the simple multi-stage algorithm being faster in both. 
These results were unexpected, as the combined algorithms were designed
for dynamic environments.
It is
worth to notice though, that
in map~1 DRRT-adv is a close second, but in map~2 it is a close last, so its
lack of reliability does not make it a good choice in this scenario. In this
environment, as in the dynamic environment, in map~1 DRRT is faster than MP-RRT,
while the opposite happens in map~2.

\begin{table}[h!]
\caption{Partially Known Environment Results, map 1.}
\label{table:partial1}
\centering
\begin{tabular}{|l||r|r|r|r@{$\ \pm\ $}l|}
\hline
\textbf{Algorithm} & \textbf{S.R.[\%]} & \textbf{C.C.} & \textbf{N.N.} &
\multicolumn{2}{c|}{\textbf{Time[s]}}\\
\hline
Multi-stage & 100 & 12204 & 1225 & 7.96 & 2.93\\
\hline
RRT-EP/N & 99 & 99076 & 1425 & 9.95 & 2.03 \\
\hline
DRRT-noadv & 100 & 37618 & 1212 & 11.66 & 15.39\\
\hline
DRRT-adv & 99 & 12131 & 967 & 8.26 & 2.5\\
\hline
MP-RRT-noadv & 99 & 49156 & 1336 & 13.82 & 17.96\\
\hline
MP-RRT-adv & 97 & 26565 & 1117 & 11.12 & 14.55\\
\hline
\end{tabular}
\end{table}

\begin{table}[h!]
\caption{Partially Known Environment Results, map 2.}
\label{table:partial2}
\centering
\begin{tabular}{|l||r|r|r|r@{$\ \pm\ $}l|}
\hline
\textbf{Algorithm} & \textbf{S.R.[\%]} & \textbf{C.C.} & \textbf{N.N.} &
\multicolumn{2}{c|}{\textbf{Time[s]}}\\
\hline
Multi-stage & 100 & 12388 & 1613 & 17.66 & 4.91\\
\hline
RRT-EP/N & 100 & 42845 & 1632 & 22.01 & 6.65 \\
\hline
DRRT-noadv & 99 & 54159 & 1281 & 32.67 & 15.25\\
\hline
DRRT-adv & 100 & 53180 & 1612 & 32.54 & 19.81\\
\hline
MP-RRT-noadv & 100 & 48289 & 1607 & 30.64 & 13.97\\
\hline
MP-RRT-adv & 100 & 38901 & 1704 & 25.71 & 12.56\\
\hline
\end{tabular}
\end{table}

\subsection{Unknown Environment Results}

Results in table~\ref{table:unknown} present the combined RRT-EP/N clearly as
the faster algorithm in unknown environments, with the multi-stage algorithm in
second place. In contrast to dynamic and partially known environments in this
same map, MP-RRT is faster than DRRT.

\begin{table}[h!]
\caption{Unknown Environment Results}
\label{table:unknown}
\centering
\begin{tabular}{|l||r|r|r|r@{$\ \pm\ $}l|}
\hline
\textbf{Algorithm} & \textbf{S.R.[\%]} & \textbf{C.C.} & \textbf{N.N.} &
\multicolumn{2}{c|}{\textbf{Time[s]}}\\
\hline
Multi-stage & 100 & 114987 & 2960 & 13.97 & 3.94\\
\hline
RRT-EP/N & 100 & 260688 & 2213 & 10.69 & 2.08 \\
\hline
DRRT-noadv & 98 & 89743 & 1943 & 18.38 & 22.01\\
\hline
DRRT-adv & 100 & 104601 & 2161 & 19.64 & 34.87\\
\hline
MP-RRT-noadv & 99 & 129785 & 1906 & 21.82 & 27.23\\
\hline
MP-RRT-adv & 100 & 52426 & 1760 & 16.05 & 10.87\\
\hline
\end{tabular}
\end{table}

\chapter{Conclusions and Future Work}
\label{sec:conclusions}

The new multi-stage algorithm proposed here has good performance in
very dynamic environments. It behaves particularly well when several small
obstacles are moving around at random. This is explained by the fact
that if the obstacles are constantly moving, they will sometimes move out of the
way by themselves, which our algorithm takes advantage of, while RRT based
ones do not, they just drop branches of the tree that could prove useful
again just a few moments later. The combined \mbox{RRT-EP/N}, although having more
operators, and automatic adjustment of the operator probabilities according to
their effectiveness, is still better than the RRT variants, but about 55\% 
slower than the
simple multi-stage algorithm. This is explained by the number of collision
checks performed, more than twice than the multi-stage algorithm, because
collision checks must be performed for the entire population, not just a single
path.

In the partially known environment, even though the difference in collision checks
is even greater than in dynamic environments, the RRT-EP/N performance is
about 25\% worse than the multi-stage algorithm. Overall, the RRT variants
are closer to the performance of both combined algorithms.

In the totally unknown environment, the combined RRT-EP/N is about 30\% faster
than the simple multi-stage algorithm, and both outperform the RRT variants,
with much smaller times and standard deviations.

All things considered, the simple multi-stage algorithm is the best choice in
most situations, with
faster and more predictable planning times, a higher success
rate, fewer collision checks performed and, above all, a much simpler
implementation than all the other algorithms compared.

This thesis shows that a multi-stage approach, using different techniques for
initial plannning and navigation, outperforms current probabilistic sampling
techniques in dynamic, partially known and unknown environments.

Part of the results presented in this thesis are published in \cite{Barriga09}.

\section{Future Work}
We propose several areas of improvement for the work presented in this thesis.
\subsection{Algorithms}
The most promising area of improvement seems to be to experiment with different on-line planners
such as a version of the
EvP (\cite{Alfaro05} and~\cite{Alfaro08}) modified to work in
continuous configuration space or a potential field navigator. Also, the local
search presented here could benefit from the use of more sophisticated
operators and the parameters for the RRT variants (such as forest size for
MP-RRT),
and the EP/N (such as population size) could benefit from being tuned specifically for this
implementation, and not simply reusing the parameters found in previous work.

Another area of research that could be tackled is extending this algorithm to
higher dimensional problems, as RRT variants
are known to work well in higher dimensions.

Finally, as RRT variants are suitable for kinodynamic planning, we only need to adapt
the on-line stage of the algorithm to have a new multi-stage planner for problems
with kinodynamic constraints.
\subsection{Framework}
The MoPa framework could benefit from the integration of a third party logic
layer, with support for arbitrary geometrical shapes, a spatial scene graph and
hierarchical maps. Some candidates would be
OgreODE~\cite{OgreODE}, Spring
RTS~\cite{SpringRTS} and
ORTS~\cite{ORTS}.

Other possible improvements are adding support for other map formats, including
discrimination of static and moving obstacles, limited
sensor range simulation and integration with external hardware such as the Lego
NXT~\cite{LegoNXT}, to run experiments in a more
realistic scenario.

\appendix


\singlespacing
\bibliographystyle{alpha}

\bibliography{../biblio}

\begin{thebibliography}{XMZT97}

\bibitem[AL02]{Atramentov02}
A.~Atramentov and S.M. LaValle.
\newblock Efficient nearest neighbor searching for motion planning.
\newblock In {\em Proceedings of the IEEE International Conference on Robotics
  and Automation}, volume~1, pages 632--637 vol.1, 2002.

\bibitem[Alf05]{Alfaro05-2}
T.~Alfaro.
\newblock Un algoritmo evolutivo para la resoluci\'on del problema de
  planificaci\'on de rutas de un robot m\'ovil.
\newblock Master's thesis, Departamento de Inform\'atica, Universidad T\'ecnica
  Federico Santa Mar\'\i a, June 2005.

\bibitem[AR05]{Alfaro05}
T.~Alfaro and M.~Riff.
\newblock An on-the-fly evolutionary algorithm for robot motion planning.
\newblock {\em Lecture Notes in Computer Science}, 3637:119, 2005.

\bibitem[AR08]{Alfaro08}
T.~Alfaro and M.~Riff.
\newblock An evolutionary navigator for autonomous agents on unknown
  large-scale environments.
\newblock {\em Intelligent Automation and Soft Computing}, 14(1):105, 2008.

\bibitem[BALS09]{Barriga09}
N.A. Barriga, M.~Araya-Lopez, and M.~Solar.
\newblock Combining a probabilistic sampling technique and simple heuristics to
  solve the dynamic path planning problem.
\newblock In {\em Proceedings XXVIII International Conference of the Chilean
  Computing Science Society (SCCC)}, 2009.

\bibitem[Ben75]{Bentley75}
J.L. Bentley.
\newblock Multidimensional binary search trees used for associative searching.
\newblock {\em Communications of the ACM}, 18(9):517, 1975.

\bibitem[BFM97]{Back97}
T.~B\"ack, DB~Fogel, and Z.~Michalewicz.
\newblock {\em {Handbook of Evolutionary Computation}}.
\newblock Taylor \& Francis, 1997.

\bibitem[BV02]{Bruce02}
J.~Bruce and M.~Veloso.
\newblock Real-time randomized path planning for robot navigation.
\newblock In {\em Proceedings of the IEEE/RSJ International Conference on
  Intelligent Robots and Systems}, volume~3, pages 2383--2388 vol.3, 2002.

\bibitem[EAA04]{Elshamli04}
A.~Elshamli, HA~Abdullah, and S.~Areibi.
\newblock Genetic algorithm for dynamic path planning.
\newblock In {\em Proceedings of the Canadian Conference on Electrical and
  Computer Engineering}, volume~2, 2004.

\bibitem[FKS06]{Ferguson06}
D.~Ferguson, N.~Kalra, and A.~Stentz.
\newblock Replanning with {RRTs}.
\newblock In {\em Proceedings of the IEEE International Conference on Robotics
  and Automation}, pages 1243--1248, 15-19, 2006.

\bibitem[HA92]{Hwang92}
Yong~K. Hwang and Narendra Ahuja.
\newblock Gross motion planning --- a survey.
\newblock {\em ACM Computing Surveys}, 24(3):219--291, 1992.

\bibitem[KL00]{Kuffner00}
Jr. Kuffner, J.J. and S.M. LaValle.
\newblock {RRT-connect}: An efficient approach to single-query path planning.
\newblock In {\em Proceedings of the IEEE International Conference on Robotics
  and Automation}, volume~2, pages 995--1001 vol.2, 2000.

\bibitem[KSLO96]{Kavraki96}
L.E. Kavraki, P.~Svestka, J.-C. Latombe, and M.H. Overmars.
\newblock Probabilistic roadmaps for path planning in high-dimensional
  configuration spaces.
\newblock {\em IEEE Transactions on Robotics and Automation}, 12(4):566--580,
  August 1996.

\bibitem[Lav98]{Lavalle98}
S.M. Lavalle.
\newblock {Rapidly-Exploring Random Trees}: A new tool for path planning.
\newblock Technical report, Computer Science Department, Iowa State University,
  1998.

\bibitem[Leg]{LegoNXT}
{Lego} {Mindstorms}.
\newblock \texttt{http://mindstorms.lego.com/}.

\bibitem[LKJ99]{Lavalle99}
S.M. LaValle and J.J. Kuffner~Jr.
\newblock Randomized kinodynamic planning.
\newblock In {\em Proceedings of the IEEE International Conference on Robotics
  and Automation}, volume~1, 1999.

\bibitem[LS02]{Li02}
Tsai-Yen Li and Yang-Chuan Shie.
\newblock An incremental learning approach to motion planning with roadmap
  management.
\newblock In {\em Proceedings of the IEEE International Conference on Robotics
  and Automation}, volume~4, pages 3411--3416 vol.4, 2002.

\bibitem[MWS07]{Martin07}
S.R. Martin, S.E. Wright, and J.W. Sheppard.
\newblock Offline and online evolutionary bi-directional {RRT} algorithms for
  efficient re-planning in dynamic environments.
\newblock In {\em Proceedings of the IEEE International Conference on
  Automation Science and Engineering}, pages 1131--1136, September 2007.

\bibitem[NG04]{Nagib04}
G.~Nagib and W.~Gharieb.
\newblock Path planning for a mobile robot using genetic algorithms.
\newblock In {\em Proceedings of the International Conference on Electrical,
  Electronic and Computer Engineering}, pages 185--189, 2004.

\bibitem[NVTK03]{Nikolos03}
I.K. Nikolos, K.P. Valavanis, N.C. Tsourveloudis, and A.N. Kostaras.
\newblock Evolutionary algorithm based offline/online path planner for {UAV}
  navigation.
\newblock {\em IEEE Transactions on Systems, Man, and Cybernetics, Part B},
  33(6):898--912, December 2003.

\bibitem[Ogr]{OgreODE}
{OgreODE}.
\newblock \texttt{http://www.ogre3d.org/wiki/index.php/OgreODE}.

\bibitem[ORT]{ORTS}
{ORTS} -- {A} free software {RTS} game engine.
\newblock \texttt{http://www.cs.ualberta.ca/\~{ }mburo/orts/}.

\bibitem[Spr]{SpringRTS}
The {Spring} {Project}.
\newblock \texttt{http://springrts.com/}.

\bibitem[Ste94]{Stentz94}
A.~Stentz.
\newblock Optimal and efficient path planning for partially-known environments.
\newblock In {\em Proceedings of the IEEE International Conference on Robotics
  and Automation}, pages 3310--3317, 1994.

\bibitem[Ste95]{Stentz95}
A.~Stentz.
\newblock The focussed {D*} algorithm for real-time replanning.
\newblock In {\em International Joint Conference on Artificial Intelligence},
  volume~14, pages 1652--1659. LAWRENCE ERLBAUM ASSOCIATES LTD, 1995.

\bibitem[TX97]{Trojanowski97}
K.M. Trojanowski and Z.J. Xiao.
\newblock Adding memory to the {Evolutionary Planner/Navigator}.
\newblock In {\em Proceedings of the IEEE International Conference on
  Evolutionary Computation}, pages 483--487, 1997.

\bibitem[Xia97]{Xiao97-2}
J.~Xiao.
\newblock {\em Handbook of Evolutionary Computation}, chapter G3.6 The
  {Evolutionary Planner/Navigator} in a Mobile Robot Environment.
\newblock IOP Publishing Ltd., Bristol, UK, UK, 1997.

\bibitem[XMZ96]{Xiao96}
J.~Xiao, Z.~Michalewicz, and L.~Zhang.
\newblock {Evolutionary Planner/Navigator}: Operator performance and
  self-tuning.
\newblock In {\em International Conference on Evolutionary Computation}, pages
  366--371, 1996.

\bibitem[XMZT97]{Xiao97}
J.~Xiao, Z.~Michalewicz, L.~Zhang, and K.~Trojanowski.
\newblock Adaptive {Evolutionary Planner/Navigator} for mobile robots.
\newblock {\em Proceedings of the IEEE Transactions on Evolutionary
  Computation}, 1(1):18--28, April 1997.

\bibitem[ZKB07]{Zucker07}
M.~Zucker, J.~Kuffner, and M.~Branicky.
\newblock Multipartite {RRTs} for rapid replanning in dynamic environments.
\newblock In {\em Proceedings of the IEEE International Conference on Robotics
  and Automation}, pages 1603--1609, April 2007.

\bibitem[ZM08]{Zhang08}
Liangjun Zhang and D.~Manocha.
\newblock An efficient retraction-based {RRT} planner.
\newblock In {\em Proceedings of the IEEE International Conference on Robotics
  and Automation}, pages 3743--3750, May 2008.

\end{thebibliography}

\end{document}